\newcommand\blfootnote[1]{%
  \begingroup
  \renewcommand\thefootnote{}\footnote{#1}%
  \addtocounter{footnote}{-1}%
  \endgroup
}

\documentclass{article}
\usepackage{ijcai26}

\usepackage{times}
\usepackage{soul}
\usepackage{url}
\usepackage[hidelinks]{hyperref}
\usepackage[utf8]{inputenc}
\usepackage[small]{caption}
\usepackage{graphicx}
\usepackage{amsmath}
\usepackage{amsthm}
\usepackage{amssymb} 
\usepackage{booktabs}
\usepackage{algorithm}
\usepackage{algorithmic}
\usepackage[switch]{lineno}
\usepackage{tikz}
\usetikzlibrary{positioning, arrows.meta, shapes.geometric, backgrounds, fit, calc}

\usepackage{tabularray}      
\usepackage{fontawesome}     
\usepackage{cleveref}        
\usepackage{mathtools}       
\usepackage{subcaption}      
\usepackage{titletoc}        
\usepackage{multirow}        
\usepackage{array}           
\usepackage{float}           

\title{Semantic Radiance Fields as Simulators for Spatial Reasoning in Real-World Scenes}

\author{
Nico Heider$^1$\and
Michał Jan Włodarczyk$^2$\and
Katarzyna Wasielewska-Michniewska$^2$\and
Przemysław Hołda$^2$\and
Martin Schieck$^1$\and
Marcin Paprzycki$^2$\and
Maria Ganzha$^2$ \And
Bogdan Franczyk $^{1,3}$ \\
\affiliations
$^1$Leipzig University\\
$^2$Systems Research Institute Polish Academy of Sciences\\
$^3$Wrocław
University of Economics\\
\emails
heider@wifa.uni-leipzig.de
}

\begin{document}

\maketitle

\begin{abstract}
     Training and evaluating spatial reasoning in embodied agents requires diverse environments that are both geometrically faithful and semantically queryable. Synthetic simulators offer ground truth semantics but sacrifice realism; simulators based on reconstructions of real-world environments have realistic appearance but lack ground truth semantics by default. We propose using Semantic Radiance Fields (SRF) as simulators for spatial reasoning agents. SRFs are a representation that unifies these requirements by lifting 2D semantic segmentations from pretrained vision models into a 3D radiance field that jointly encodes geometry, appearance, and per-class semantic identity. The resulting fields are reconstructed from posed RGB captures of real scenes and support novel-view synthesis, semantic and free-space queries within a single grounded representation. This enables the efficient generation of diverse real-world environments to train and evaluate spatial reasoning models. As an example application, we outline an SRF-driven simulator for an orchard apple-reaching task, in which the radiance field supplies camera rendering, semantic ground truth, and occupancy queries to a physics engine.
     
\end{abstract}

\blfootnote{\hspace{-1.8em}Accepted at the IJCAI 2026 Workshop on Spatio-Temporal Reasoning and Learning (STRL), oral presentation.}

\section{Introduction}
Representing the diversity of real-world environments is a central bottleneck for training embodied AI systems~\cite{deitke2023phone2proc}. Agents that navigate, manipulate, or answer questions about their surroundings must reason not only about what objects are present, but about where they are, what occludes them, and how the scene would appear from unseen viewpoints. Progress on these problems depends on the environments in which reasoning systems are trained and evaluated. Two paradigms currently dominate to create a large amount of diverse training data to train these agents. \emph{Synthetic procedural simulators} ~\cite{duan2022survey} offer programmatic control, ground-truth supervision, and scene perturbations, but sacrifice the visual and geometric appearance of natural scenes. \emph{Generative simulators} ~\cite{brooks2024sora,bruce2024genie,yang2024unisim} can produce large amounts of training data and, given sufficient training, cover a wide range of visual scenes. However, they do not guarantee multiview consistency, and do not natively expose a persistent, queryable 3D state. Neural scene representations such as radiance fields (NeRFs)~\cite{mildenhall2020nerf} enable photorealistic reconstruction of real environments from posed images, and recent extensions lift 2D segmentations into the field to produce per-class semantic outputs~\cite{zhi2021semanticnerf,meyer2024fruitnerf}. These semantic radiance fields have primarily been developed as tools for reconstruction and querying and have so far mainly been used as RL training environments for locomotion ~\cite{byravan2023nerf2real,zhou2024splatgym}, but not for tasks that require the agent to reason about object identity. We argue that a semantic radiance field, exposing rendered views, per-class labels, and free-space queries, is sufficient to train a vision-based RL agent directly in reconstructions of real scenes, bridging procedural and reconstruction-based simulators.

\section{Semantic Radiance Fields}
\begin{figure*}[t]
\centering
\includegraphics[width=\textwidth]{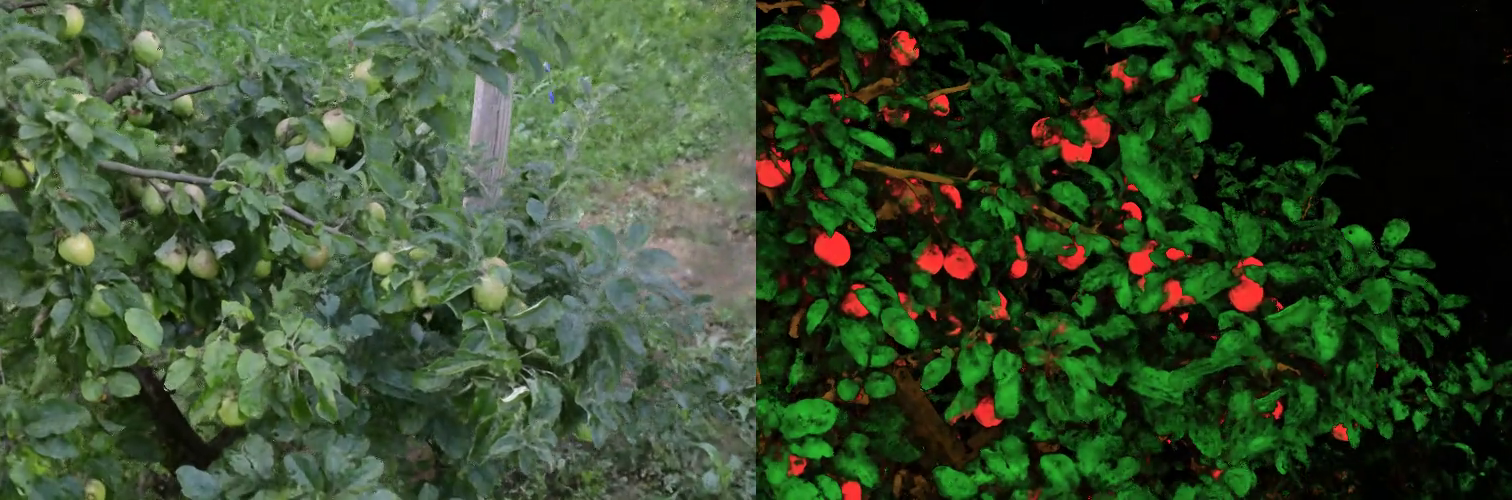}
\caption{Radiance Field rendering (left) and Semantic Field blend rendering (right) from a trained Semantic Radiance Field. Classes for the SRF blend rendering: apple, branch, leaf; colored to match the class for visualization.}
\label{fig:placeholder}
\end{figure*}

We extend the FruitNeRF formulation~\cite{meyer2024fruitnerf} from a single semantic channel to $C$ class channels, yielding a representation in which any subset of the scene can be queried by semantic identities. Our pipeline takes an unordered set of posed RGB images $\{I_i\}_{i=1}^N$ of a real scene, generates per-image multi-class segmentation masks with a pretrained vision model, and jointly optimizes a radiance field whose $C$ semantic heads each predict an independent binary probability for their respective class at every 3-D point.

\subsection{Volumetric Rendering}
A neural radiance field~\cite{mildenhall2020nerf} represents a scene as a continuous function mapping a 3D position $\mathbf{x} \in \mathbb{R}^3$ and view direction $\mathbf{d} \in \mathbb{S}^2$ to a density $\sigma \in \mathbb{R}_{\geq 0}$ and an emitted RGB radiance $\mathbf{c} \in [0,1]^3$. We adopt the Nerfacto~\cite{tancik2023nerfstudio} factorization, in which a density field $\mathcal{F}_\sigma : \mathbf{x} \to (\sigma, \mathbf{h})$ produces a latent feature $\mathbf{h}$ alongside the density, and an appearance field $\mathcal{F}_\mathbf{c} : (\mathbf{h}, \mathbf{d}) \to \mathbf{c}$ produces color. The expected color of a pixel is computed by alpha-compositing $K$ samples along a ray $\mathbf{r}(t) = \mathbf{o} + t\mathbf{d}$:
\begin{equation}
\hat{\mathbf{C}}(\mathbf{r}) = \sum_{k=1}^{K} \hat{T}(t_k)\,\alpha(\sigma(t_k)\delta_k)\,\mathbf{c}(t_k),
\label{eq:render_color}
\end{equation}
with $\hat{T}(t_k) = \exp\!\left(-\sum_{a=1}^{k-1}\sigma(t_a)\delta_a\right)$, $\delta_k = t_{k+1} - t_k$, and $\alpha(x) = 1 - \exp(-x)$.

\subsection{Multi-Class Semantic Field}
\label{sec:method:semantic}

To encode semantics, we attach $C$ independent binary semantic heads to the shared density backbone:
\begin{equation}
\mathcal{F}_s : \mathbf{h} \to \mathbf{s} \in \mathbb{R}^C,
\end{equation}
where each output $s_c$ is an unconstrained logit for class $c$.
Unlike SemanticNeRF~\cite{zhi2021semanticnerf}, which uses a single softmax head that enforces mutual exclusion across classes, our $C$ heads are independent: each is activated by a sigmoid, so a point may simultaneously belong to multiple classes.
Following~\cite{zhi2021semanticnerf,meyer2024fruitnerf}, the semantic head depends only on position, semantic identity is view-independent, and gradients from the semantic loss are not propagated back through the density field, preventing the geometry from collapsing onto class boundaries. The expected per-class logits at a pixel are obtained by the same volumetric accumulation used for color:
\begin{equation}
\hat{\mathbf{S}}(\mathbf{r}) = \sum_{k=1}^{K} \hat{T}(t_k)\,\alpha(\sigma(t_k)\delta_k)\,\mathbf{s}(t_k).
\label{eq:render_sem}
\end{equation}

\subsection{Lifting 2D Segmentations to 3D}
\label{sec:method:lifting}

We do not assume hand-labeled semantic ground truth. For each input image $I_i$ we obtain a multi-class segmentation $M_i : \Omega \to \{0,1\}^C$ from a pretrained model. In our experiments, we use SAM 3~\cite{carion2025sam3segmentconcepts} queried with a fixed vocabulary of class prompts. Letting $y_c(\mathbf{r}) \in \{0,1\}$ denote the binary ground-truth label class for the pixel through which ray $\mathbf{r}$ passes, the semantic loss applies independent binary cross-entropy to each class:
\begin{equation}
  \mathcal{L}_{\text{sem}} = \frac{1}{|\mathcal{R}|} \sum_{\mathbf{r} \in
  \mathcal{R}}
    \operatorname{BCE}\bigl(\sigma(\hat{\mathbf{S}}(\mathbf{r})),\,
  \mathbf{y}(\mathbf{r})\bigr),
  \label{eq:sem_loss}
  \end{equation}
  where $\operatorname{BCE}(\hat{\mathbf{p}}, \mathbf{y}) =
  -\sum_{c=1}^{C}\bigl[y_c \log \hat{p}_c + (1-y_c)\log(1-\hat{p}_c)\bigr]$ and
  $\sigma$ is the sigmoid applied element-wise.
  The total training objective combines this with the standard photometric loss:
  \begin{equation}
  \mathcal{L} = \mathcal{L}_{\text{photo}} + \lambda \mathcal{L}_{\text{sem}},
  \quad
  \mathcal{L}_{\text{photo}} = \frac{1}{|\mathcal{R}|}
  \sum_{\mathbf{r}\in\mathcal{R}} \|\mathbf{C}(\mathbf{r}) -
  \hat{\mathbf{C}}(\mathbf{r})\|_2^2.
  \label{eq:total_loss}
\end{equation}
We set $\lambda = 1$ in all experiments.

\subsection{Query Interface}
\label{sec:method:queries}

A trained SRF exposes three operations that downstream reasoning agents can call:
\begin{itemize}
    \item \textbf{Render}$(\mathbf{P})$: given a camera pose $\mathbf{P} \in \mathrm{SE}(3)$, produce a posed RGB image, a semantic map, and a depth map by evaluating Eq.~\ref{eq:render_color} and Eq.~\ref{eq:render_sem} over the corresponding rays.
    \item \textbf{Semantic}$(\mathbf{x})$: return the per-class probability $\sigma(\mathbf{s}(\mathbf{x})) \in [0,1]^C$ at any 3D point.
    \item \textbf{Occupancy}$(\mathbf{x})$: return $\sigma(\mathbf{x})$, for collision detection.
\end{itemize}
Together these support the viewpoint sampling, object localization, collision-detection, and the per-pixel ground-truth supervision.

\section{Example Application: Apple-Reaching in an Orchard SRF}
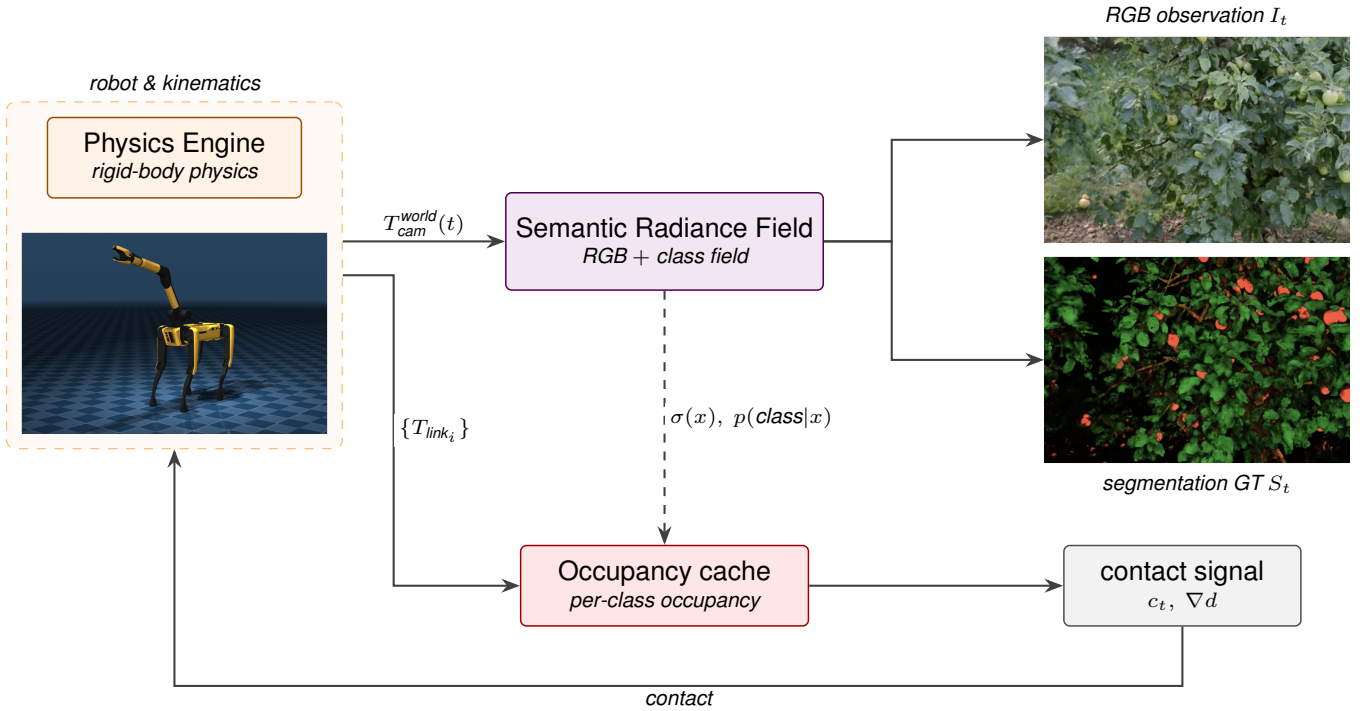
\begin{figure*}[t]
  \centering
  \resizebox{\textwidth}{!}{%
  \begin{tikzpicture}[
    font=\sffamily\small,
    >={Stealth[length=2.2mm]},
    module/.style 2 args={
      draw=#1!60!black, fill=#1!10, line width=0.5pt,
      rounded corners=2pt, align=center,
      minimum width=3.0cm, minimum height=0.95cm,
      font=\sffamily\small
    },
    img/.style={draw=black!30, line width=0.4pt, inner sep=0pt},
    flow/.style={->, line width=0.7pt, draw=black!75},
    ann/.style={font=\sffamily\scriptsize\itshape,
                fill=white, inner sep=1.5pt}
  ]

  \node[module={orange}{}] (mj) {Physics Engine\\[-1pt]\scriptsize\itshape rigid-body physics};
\node[img, below=4mm of mj] (spot)
  {\includegraphics[width=3.6cm]{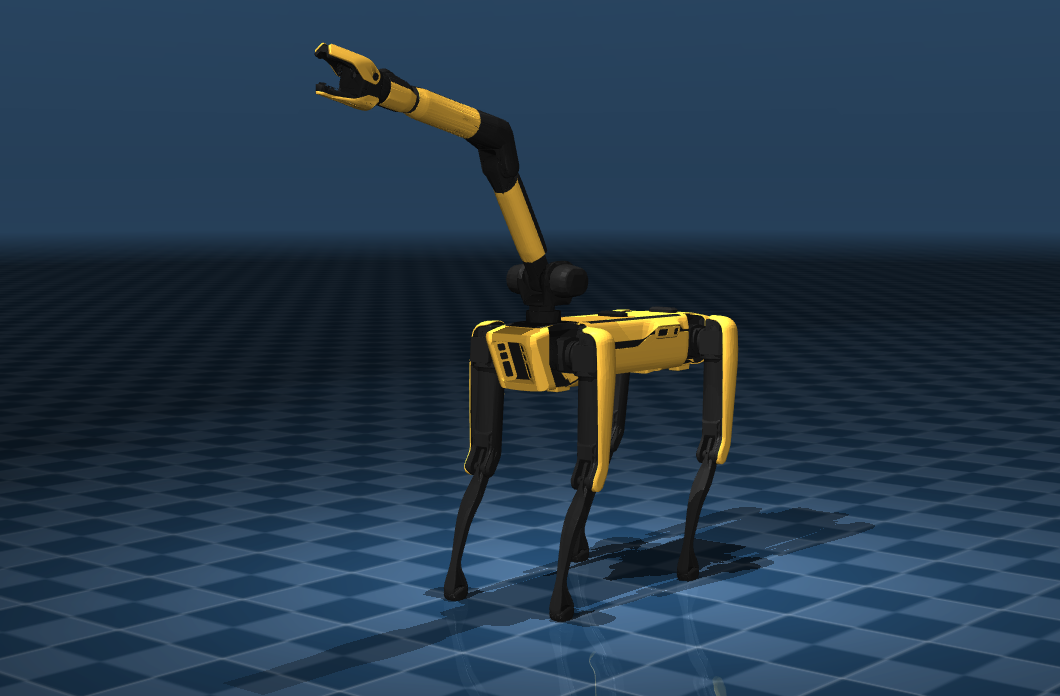}};
\coordinate[below=0mm of spot] (spotpad);
\begin{pgfonlayer}{background}
  \node[draw=orange!55, dashed, rounded corners=3pt,
        fit=(mj)(spot)(spotpad), inner sep=5pt, fill=orange!4] (mjbox) {};
\end{pgfonlayer}
\node[font=\sffamily\scriptsize\itshape, above=0pt of mjbox.north]
  {robot \& kinematics};

  \node[module={violet}{},
        minimum width=3.4cm, minimum height=1.15cm]
        at ($(mjbox.east)+(38mm,4mm)$) (srf)
    {Semantic Radiance Field\\[-1pt]\scriptsize\itshape RGB $+$ class field};

  \node[module={red}{}, below=30mm of srf, minimum width=3.4cm] (voxel)
    {Occupancy cache\\[-1pt]\scriptsize\itshape per-class occupancy};

  \node[img, right=26mm of srf, yshift=12mm] (rgb)
    {\includegraphics[width=3.6cm]{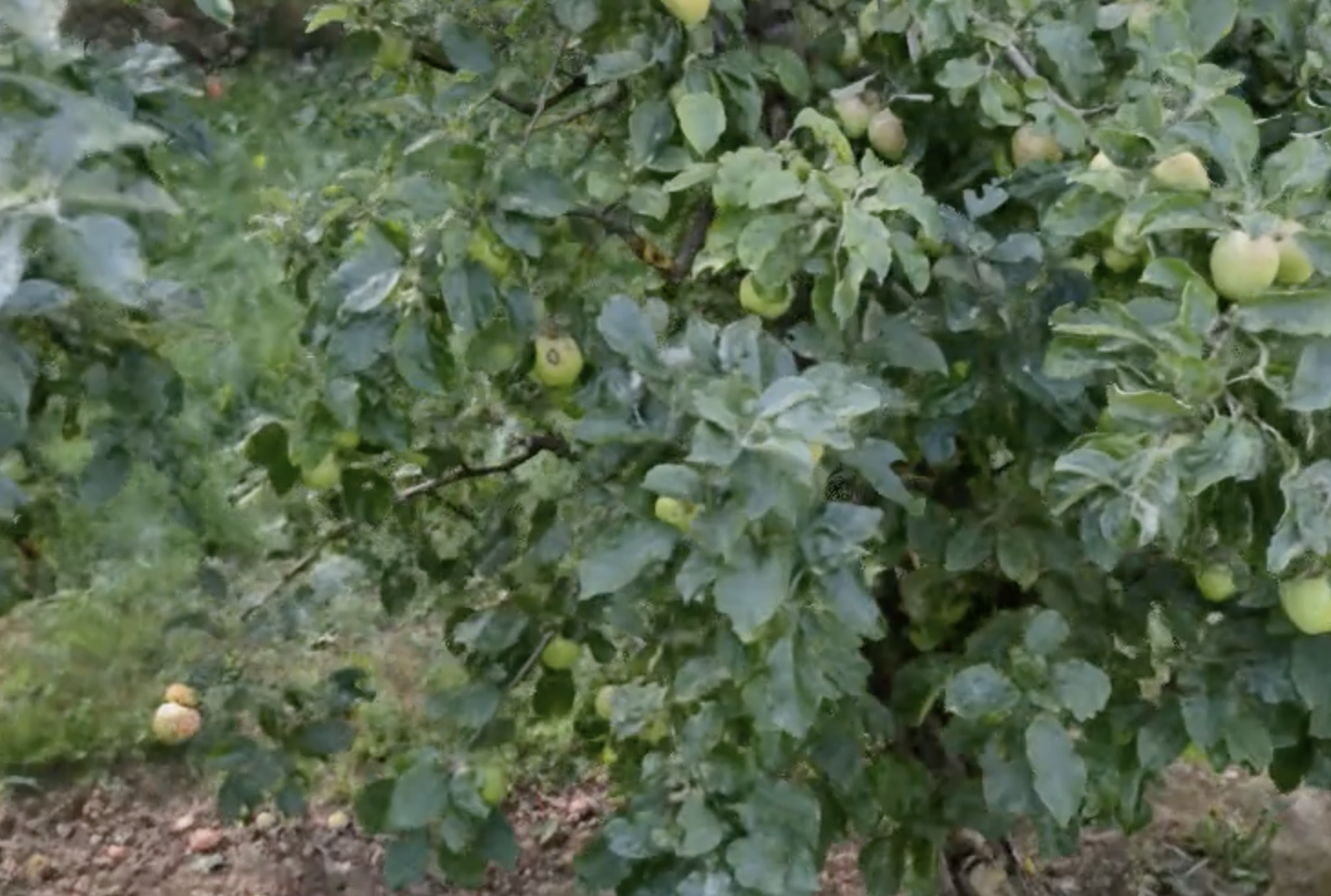}};
  \node[ann, above=0.8mm of rgb] {RGB observation $I_t$};

  \node[img, right=26mm of srf, yshift=-14mm] (seg)
    {\includegraphics[width=3.6cm]{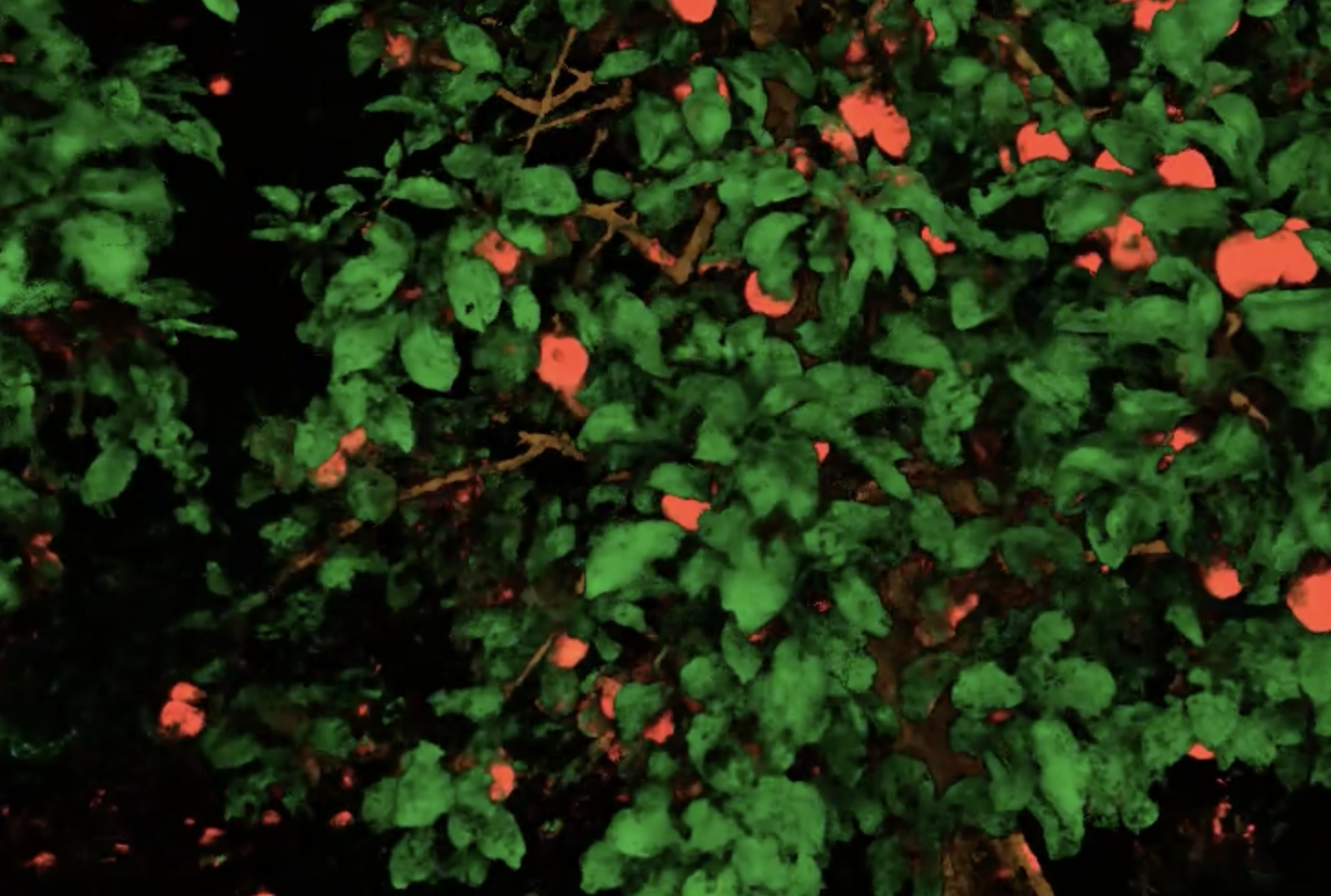}};
  \node[ann, below=0.8mm of seg] {segmentation GT $S_t$};

  \node[module={gray}{}, right=30mm of voxel, anchor=west,
        minimum width=2.8cm] (col)
    {contact signal\\[-1pt]\scriptsize\itshape $c_t,\ \nabla d$};

  \draw[flow] (mjbox.east|-srf) -- 
    node[ann, above]{$T^{\text{world}}_{\text{cam}}(t)$} (srf.west);


  \draw[flow] (mjbox.east) -- ++(6mm,0) -- ++(0,0) |-
  node[ann, pos=0.25, right]{$\{T_{\text{link}_i}\}$} (voxel.west);

 \draw[flow] (srf.east) -- ++(8mm,0) |- (rgb.west);
 \draw[flow] (srf.east) -- ++(8mm,0) |- (seg.west);

  \draw[flow, dashed] (srf.south) -- 
    node[ann, right]{$\sigma(x),\ p(\text{class}|x)$} (voxel.north);

  \draw[flow] (voxel.east) -- (col.west);

  \draw[flow] (col.south) -- ++(0,-7mm) -|
    node[ann, pos=0.25, below]{contact} (mjbox.south);

  \end{tikzpicture}%
  }
  \caption{Semantic Radiance Field as a simulator. A physics engine  (e.g., MuJoCo) handles rigid-body 
  dynamics and exposes the camera pose $T^{\text{world}}_{\text{cam}}(t)$ and 
  link transforms $\{T_{\text{link}_i}\}$ each step. The SRF renders an RGB 
  observation $I_t$ and a semantic map $S_t$ from the camera pose; its density and per-class probabilities can be distilled offline into an occupancy cache (e.g., a voxel grid or octree) used for 
  collision and goal-reaching queries. Contact signals are returned to the physics engine to close the loop.}
  \label{fig:pipeline}
\end{figure*}

As an example application of our pipeline, we show how SRFs can serve embodied agent training by acting as the camera renderer and segmentation mask provider for an apple-reaching policy, where the goal of a robot is to bring its manipulator end-effector close to a target apple.

\subsection{Setup}

\paragraph{Scenes.} We reconstruct an apple tree scene from
FruitNeRF~\cite{meyer2024fruitnerf}, captured with 311 posed RGB images at $6000
\times 4000$\,px. We use the provided camera poses without refinement. Per-image
segmentation masks are generated by SAM~3~\cite{carion2025sam3segmentconcepts}
queried independently with three text prompts (\emph{apple}, \emph{branch}, and
\emph{leaf}) producing one binary mask per class per frame. These masks are composited into a single label image per frame (background\,=\,0, apple\,=\,1, branch\,=\,2, leaf\,=\,3) and passed to the NeRF training pipeline as semantic supervision.

\paragraph{SRF Training.} We build on FruitNeRF~\cite{meyer2024fruitnerf} with the $C$ independent binary semantic heads described in Sec.~\ref{sec:method:semantic}. The model is trained for 500\,000 iterations with a batch size of 4\,096 rays, using the Adam optimiser with an initial learning rate of $10^{-2}$ decayed exponentially, and mixed-precision arithmetic. Input images are downscaled by a factor of 4 (to $1500 \times 1000$\,px) before training. Each scene trains in approximately 4 hours on a single NVIDIA H100 GPU.

\paragraph{Simulator Setup.} 
The SRF can be distilled offline into an occupancy cache (e.g., a voxel grid or octree) to efficiently query contact information for the simulator and to provide reward information for the policy. A physics engine such as MuJoCo handles the rigid-body dynamics and exposes the pose of the robot's wrist camera and the link transforms at each step. The SRF serves as the renderer, providing RGB observations and a semantic map from the camera pose. Contact signals returned to the physics engine close the loop.

\paragraph{Task and Agent.} The reward formulation is designed to encourage the agent to approach target objects (fruits), whose positions are obtained from the semantic occupancy cache. To promote collision-free kinematics, the system utilizes the derived contact signal to terminate episodes upon detecting collisions with designated avoidance classes, such as branches. A full training study is beyond the scope of this work; the example specifies the set of signals (observations, semantic ground truth, and rewards) that an SRF reconstructed from captures of a real scene can supply to close the simulation loop.

\section{Conclusion}

We presented Semantic Radiance Fields, a representation that lifts multi-class 2D segmentations into a radiance field to produce queryable, photorealistic reconstructions of real-world scenes as suitable training and evaluation environments for spatial reasoning agents. By extending real-world geometry with discrete semantic structure, SRFs can serve as drop-in simulators for any task that requires novel view synthesis with ground truth semantics. Our example application illustrates that a single SRF can serve simultaneously as a renderer, a segmentation oracle, and a collision detector for embodied agents. Several extensions are immediate. The same lifting procedure applied to 3D Gaussian Splatting~\cite{kerbl2023gaussiansplatting}  would reduce training time and enable real-time rollouts. Dynamic SRFs, in which a temporal axis is added to the semantic field, would extend the SRF from spatial to spatio-temporal reasoning, a natural next step for the present work.

\bibliographystyle{named}
\bibliography{ijcai26}

\end{document}